# A Lightweight and Accurate Face Detection Algorithm Based on Retinaface

Baozhu Liu , Hewei Yu

School of Computer Science and Engineering

South China University of Technology

Guangzhou, 510006, P.R.China

583049537@qq.com, hwyu@scut.edu.cn

ABSTRACT. *In this paper, we propose a lightweight and accurate face detection algorithm LAFD (Light and accurate face detection) based on Retinaface. Backbone network in the algorithm is a modified MobileNetV3 network which adjusts the size of the convolution kernel, the channel expansion multiplier of the inverted residuals block and the use of the SE attention mechanism. Deformable convolution network(DCN) is introduced in the context module and the algorithm uses focal loss function instead of cross-entropy loss function as the classification loss function of the model. The test results on the WIDERFACE dataset indicate that the average accuracy of LAFD is 94.1%, 92.2% and 82.1% for the "easy", "medium" and "hard" validation subsets respectively with an improvement of 3.4%, 4.0% and 8.3% compared to Retinaface and 3.1%, 4.1% and 4.1% higher than the well-performing lightweight model, LFFD. If the input image is pre-processed and scaled to 1560px in length or 1200px in width, the model achieves an average accuracy of 86.2% on the 'hard' validation subset. The model is lightweight, with a size of only 10.2MB.*
**Keywords:** face detection; lightweight network; Retinaface; deformable convolution; focal loss function.

1. **Introduction.** Face recognition is widely used in people's daily life. The face recognition mentioned in this paper is not for the recognition of individual faces, but refers to localization of faces in pictures or videos and counting of faces. The development of face detection algorithms can be divided into three phases, namely the early algorithms, the Adaptive Boosting framework [1] , and the deep learning era.

Early face recognition used a modular matching technique, which involves using a template image of a face to match various locations in the detection image to determine if there is a face at that location. A representative work was the algorithm proposed by Rowley (Neural network-based face detection[2] ), which used a 20x20 dataset to train a Multi-layer Perceptron [3] model with good accuracy, but ran slowly.

In 1997, Margineantu et al. proposed a face recognition algorithm in the AdaBoost framework. The boost algorithm is an ensemble learning algorithm based on PAC (probably approximately correct) learning theory. In 2001, Viola and Jones designed a face detection algorithm [4] It used simple Haar-like [5] features and cascaded AdaBoost classifiers to construct a detector that improved detection speed by two orders of magnitude over previous methods and maintained good accuracy. This approach is known as the VJ framework. the VJ framework is a landmark achievement in the history of face detection and laid the foundation for the AdaBoost-based target detection framework.

With the development of deep learning techniques, deep neural networks began to be used in face recognition, such as Convolutional Neural Networks (CNN) with tens or even hundreds of layers [6] . The application of deep networks enables models to greatly optimize the extraction of face features.

In 2017, Shuo Yang, Ping Luo et al. proposed Faceness-Net which is a face detection algorithm based on deep convolutional networks [7] . Firstly, the algorithm performs the detection of local features of faces, scores each part of the face (nose, eyes, etc.) using multiple classifiers based on deep convolutional networks; then analyzes the most likely face regions based on the score of each part; finally trains a CNN for face classification and boundary regression to further enhance its effectiveness. When the Faceness-Net model had 100 false positives samples on the FDDB dataset, the face detection rate was close to 88%, an improvement of almost 4 percentage points compared to the best detection accuracy in academia at the time

In 2022, the YOLOv7 model was proposed by Bochkovskiy et al. Lightweight models built on top of YOLOv7, such as yolov7-tiny[8] , achieved average accuracies of 94.7%, 92.6% and 82.1% on easy, medium and hard subsets of the WIDERFACE validation set, with a model size of 11.8 MB; yolov7-lite-s [8] , achieved average accuracies of 92.7%, 89.9% and 78.5% on three subsets of the validation set, with a model size of only 2.3MB.

To enable real-time detection on devices with lower computational costs, in May 2020, Imperial, Middlesex University London, and InsightFace jointly proposed RetinaFace, which is fast in detection and better meets the requirements of face detection speed in practical applications. However, the accuracy of face detection in this network is low, the face location is not accurate, and the detection performance is especially poor for complex faces such as small faces. To solve this problem, this paper proposes a Light and Accurate Face Detection (LAFD) method based on RetinaFace with the following algorithm improvements:

(1) Using the optimized backbone network of MobileNetV3 [9] , we fine-tune the size of the convolutional kernel, the SE attention mechanism (Squeeze-and-Excitation Networks) [10] and the expansion multiplier of the number of feature map channels in the inverted residuals block [11] in the MobileNetV3 backbone network, so that the model focuses more on facial features.

(2) Introducing DCN [12] (deformable convolution network) in the enhanced feature layer, replacing the convolution kernel size of 3x3 with DCN to enhance its feature extraction capability and experimentally demonstrating the effectiveness of the improvement.

(3) Optimization of loss function. Using the Focal Loss Function [13] to replace the cross-entropy loss has a mitigating effect on the problem that individual samples are difficult to identify and most of prior boxes fail to detect faces. The function can balance the training samples of faces and non-faces and reduce the loss of easily trained correct samples, and the experimental results are compared to verify the effectiveness of the model using the Focal Loss Function.

(4) Pre-processing of the input images. After several experiments, the model works best on the WIDERFACE validation set by expanding the input image size to a maximum length of 1560 or a maximum width of 1200 proportionally.

2. **Related Work.** RetinaFace is a single-stage, lightweight face detection network that achieves 90.70%, 88.16% and 73.82% on the "easy", "medium" and "hard" validation subsets of the WIDERFACE dataset respectively when using MobileNetV1 [14] as the backbone network.

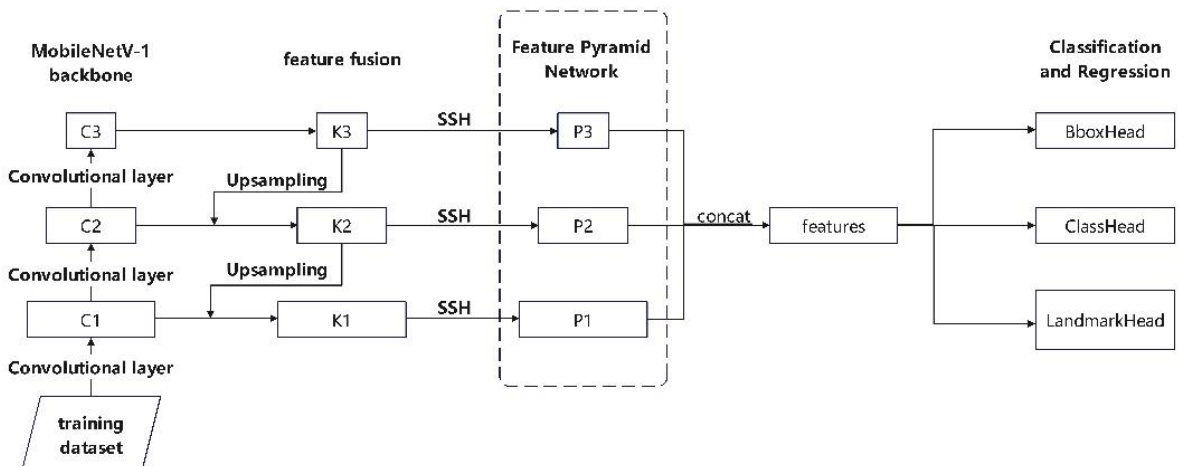

FIGURE 1. The main process of the Retinaface algorithm.

The main process of the RetinaFace algorithm is shown in Fig. 1. Firstly, input the training dataset into MoblieNetV-1 backbone. The outputs of the convolutional layer are noted as $C_1, C_2, C_3$. The outputs of the three convolutional layers here do not mean that there are only three convolutional layers in backbone network, but the outputs of the three selected convolutional layers in the backbone network. After extracting the feature maps, the feature fusion is performed on these feature maps, using the method of up-sampling (bilinear interpolation) to ensure that the two adjacent layers are of the same size, and then matrix summing the two layers to obtain $K_1, K_2, K_3$, forming the feature pyramid structure for this step, and then using the feature maps of each selected layer processed as input into the context module to obtain $P_1, P_2, P_3$. Finally, input $P_1, P_2, P_3$ into the classification and regression branch to obtain prediction results.

Each feature layer in the feature pyramid aggregates face information at different scales. As small faces are not easily detected in face detection, Retinaface uses SSH (Single Stage Headless Face Detector) [15] as the context module (Fig. 2) to enhance the receptive field of the model and improve the detection rate of small faces.

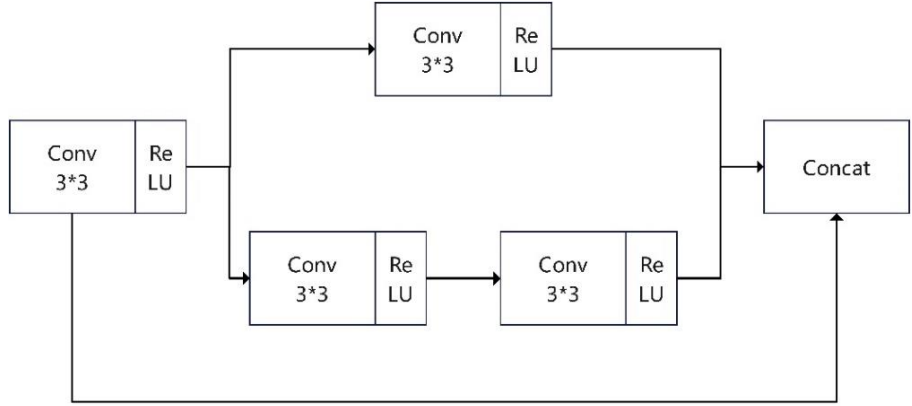

FIGURE 2. The SSH context module is three parallel convolutions, using one, two and three 3*3 convolutions respectively.

In Fig. 2, two 3*3 convolutions are simulating the effect of 5*5 convolution, and three 3*3 convolutions are simulating the effect of 7*7 convolution. The different sizes of convolution kernels provide different receptive fields for feature extraction and fully extract image features.

In the RetinaFace, the authors pre-designed a lot of prior boxes of the face in the $P_1, P_2, P_3$ so that the face at any position of the image can be detected. The model presupposes that each pixel in the feature map corresponds to two sizes of prior boxes at the location in the original image, so the number of prior boxes in the 8*8 size feature map is 8*8*2= 128 prior boxes, which corresponds to 64 different locations of face prior boxes of two sizes in the original image. The size of the prior boxes is a hyperparameter.

Four types of Head prediction are used in Retinaface, namely face classification prediction, face box point regression, face key point regression, and 3D dense point regression.

For each prior box, a binary classification prediction of whether it contains a face is performed, and the number of feature channels obtained from SSH is adjusted to the number of prior boxes*2 using a 1*1 convolution, with the two parameters representing the probability of a human face and an unarmed face respectively, and finally a cross-entropy function is used as the loss function.

The face box point regression is intended to calculate the center offset and length-width offset needed for a prior box to accurately wrap the face. $1*1$ convolution is used in Retinaface to adjust the number of feature map channels to the number of prior boxes*4, and the four parameters represent one prior box center x-axis offset, center y-axis offset, edge length x-axis offset, and edge length y-axis offset, and finally Smooth-L1(loss function used in face-r-cnn [19] ) as the loss function. The loss function is given as:

$$Smooth\ L1 = \begin{cases} 0.5x^2, & where\ |x| < 1 \\ |x| - 0.5, & where\ x < -1\ or\ x > 1 \end{cases}$$

(1)

Retinaface set the left and right eyes, both corners of the mouth and the nose of a person as the key points of the face which is an additional training task. The channels are adjusted to 10*number of prior boxes by using 1*1 convolution, with each prior box having ten
4

parameters representing the x and y axis offsets of key points. The loss function used is still Smooth-L1.

When setting prior boxes in the model, two different sizes of prior boxes are set for each pixel of the feature map extracted by SSH, resulting in the existence of a large number of prior boxes. The final model result may show multiple overlapping face boxes around the face. To solve this problem, Retinaface uses NMS (non-maximum suppression) to find the most suitable face box.

3. **Light and Accurate Face Detection (LAFD).** The LAFD optimizes the model by making three primary improvements: the backbone network, the context module, and the loss function.

3.1. **Backbone Network Improved from MobilenetV3.** The channel expansion in the backbone module of MobileNetV3 is small, which is not conducive to the extraction of sufficient image information. Therefore, we appropriately increase the channel expansion multiplier in the inverted residuals module and set the multiplier to 6 for the convolutional layers with same output channels to fully extract the image information and set the value to 3 for the convolutional layers with increased output channels to carry out the transition of expanding the channels of the feature map. The convolutional kernels of the MobileNetV3 network size are either 3 or 5, and their receptive fields are limited. To improve the accuracy, our improved backbone network introduces $7*7$ convolutional kernels and reduces consecutive convolutional layers that have the same convolutional kernel size, so that the model has a richer receptive field. The MobileNetV3 network only uses SE attention mechanism in the convolutional layer with 40 output feature map channels in the early stage, and the later convolutional layers of the network all use the SE attention mechanism. In this paper, we add the SE attention mechanism in the convolutional layers with the input channel number of 24,80 and reduce the use of mechanism in the convolutional layers with the input channel number of 40,112. The improved network uses the SE attention mechanism in the early stage to enhance the feature extraction capability of the model and reduces the use of the SE attention mechanism in the late stages of the network to increase the randomness of the model training and alleviate the problem of the training model staying at the extreme points.

The structure of improved backbone network is shown in Table 1.

TABLE 1. Backbone Network Improved from MobilenetV3

| Expansion* | out_c | k_size* | stride | RE/HS | SE* | role |
|---|---|---|---|---|---|---|
| - | 16 | 3 | 2 | HS | 0 | - |
| 1 | 16 | 3 | 1 | RE | 0 | - |
| 6* | 24 | 5* | 2 | RE | 0 | - |
| 3 | 24 | 7* | 1 | RE | 1* | - |
| 6* | 40 | 3* | 2 | RE | 1 | - |
| 6* | 40 | 3* | 1 | RE | 0* | - |
| 3 | 40 | 5 | 1 | RE | 1 | *Stage1* |
| 6* | 80 | 7* | 2 | HS | 1* | - |
| 6* | 80 | 3 | 1 | HS | 0 | - |
| 6* | 80 | 3 | 1 | HS | 1* | - |
| 3* | 80 | 5* | 1 | HS | 1* | - |
| 6* | 112 | 7* | 1 | HS | 0* | - |
| 3* | 112 | 7* | 1 | HS | 1 | *Stage2* |
| 6* | 160 | 5 | 2 | HS | 1 | - |
| 6 | 160 | 5 | 1 | HS | 1 | - |
| 3* | 160 | 5 | 1 | HS | 1 | *Stage3* |
| - | 960 | 1 | 1 | HS | 0 | *Conv2d 1\*1* |
| - | 960 | - | 1 | - | 0 | *Avg pool* |
| - | 1280 | 1 | 1 | HS | 0 | *NBN* |
| - | *class_num* | 1 | 1 | HS | 0 | *NBN* |

Note: Column indexes indicate the channel expansion multiplier in the inverted residuals block, the number of output channels per layer, the convolution kernel size, the stride, the use of ReLU/H-Swish for the activation function, the use of SE Module or not, and the role. Stage1,2,3 represent the three layers of feature images selected by the feature pyramid (Fig.1 $C_1, C_2, C_3$), NBN indicates a $1*1$ convolution without the BatchNorm layer. The sign * indicates that the backbone network has been adjusted relative to the MobileNetV3 network.

3.2. **Deformable Convolutional Networks.** The normal convolution operation uses a rectangular convolution kernel to perform a sliding window on the feature map to calculate the weights during learning, but the targets of interest are often irregularly shaped, in which case rectangle-like feature extraction can lead to scattering of the learned image features. In target detection tasks, a challenge in recognizing content is to identify pose, scale, and geometric variations of the target, and the concept of DCN (Deformable Convolution Network) [12] is introduced here.

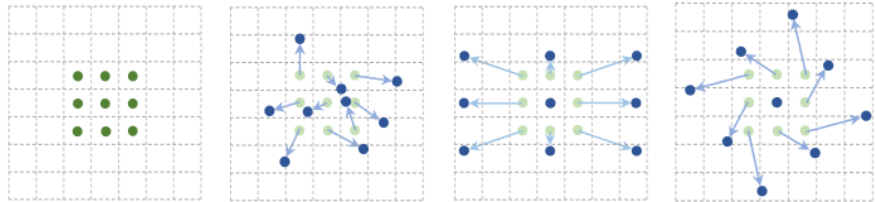

FIGURE 3. [12] Normal convolution sampling position a) and deformable convolution sampling positions b), c), d)

As shown in Fig. 3, after replacing the convolution with DCN convolution, each sample location point is offset by a certain degree on the x and y axes, thus allowing the model to better learn the features of the picture irregular objects, and the way the model adapts depends on the input feature image.

The flow of DCN is as follows:

(1) The original image batch (size $b*h*w*c$) is subjected to a normal convolution, and the output image batch (size $b*h*w*2c$) is output with an output channel of $2c$ and the offset $(\Delta x, \Delta y)$ of each pixel in the original image batch. Where $b$ denotes the number of training batches, $h$ denotes the height of the image, $w$ denotes the width of the image, and $c$ denotes the number of input channels.

(2) The pixel is corrected according to the offset, and let the input feature map be P. The corrected pixel value $P^`$ is
$$P^`(x,y) = P(x + \Delta x, y + \Delta y) \tag{2}$$

(3) Computation of float pixels using bilinear interpolation. For example, the original image pixel point $p_{2,2}$ represents the value of the pixel at position (2,2) of the original image. If the convolution process yields $p_{2,2}$ offset by -0.8 on the x-axis and 1.2 on the y-axis, the modified pixel $p'_{2,2}$ is the pixel value at position (1.2,3.2). Due to the presence of float pixels, the pixel values at position (1.2,3.2) are calculated using bilinear interpolation of the values of $p_{13}, p_{23}, p_{14}, p_{24}$.

(4) The output feature image is obtained by using ordinary convolution.

In this paper, all convolutions in the SSH context module with a convolution kernel size of $3x3$ are replaced with a deformable convolution with a convolution kernel size of $3x3$

3.3. **Focal Loss Function.** The cross-entropy loss function Retinaface used:
$$L = \frac{1}{N}\sum_i L_i = \frac{1}{N}\sum_i -[y_i * \log(p_i) + (1 - y_i) * \log(1 - p_i)] \tag{3}$$
$y_i$ is the label value of the $i$-th sample, as 0 or 1, $p_i$ is the prediction probability of the model for the $i$-th sample as a positive sample, with the value of [0,1]. If the sample is a positive sample (face, $y_i = 1$), the Cross Entropy Loss is
$$CE = -\log(p_i) \tag{4}$$
If the sample is negative (background, $y_i = 0$), the Cross Entropy loss is
$$CE = -\log(1 - p_i) \tag{5}$$
To solve the problem of sample imbalance and hard-to-classify samples, the Focal Loss Function is introduced and the prediction accuracy $p_t$ is defined as
$$p_t = \begin{cases} p & ,where\ y = 1 \\ 1 - p, & otherwise \end{cases} \tag{6}$$
For the sample imbalance problem, introduce a weighting factor $\alpha$ in the range [0,1]. When $y_i = 1$, the training sample $i$ is a positive sample with weight $\alpha$, when $y_i = 0$, the training sample $i$ is a negative sample with weight $(1 - \alpha)$. Define the weighting formula is $\alpha_t$, the Cross-Entropy Loss which is introduced $\alpha_t$ is changed to
$$CE(p_t, \alpha_t) = -\alpha_t \log(p_t) \tag{7}$$
The weighting factor α alleviates the problem of unbalanced sample data by adjusting the proportion of positive and negative samples in the Cross-Entropy Loss.

To address the problem of distinguishing "hard" samples, a moderating factor $\gamma$ is introduced into the Cross-Entropy Loss function in the range $[0, +\infty)$. The function after introduction is

$$CE(p_t, \gamma) = -(1-p_t)^\gamma \log(p_t) \tag{8}$$

Moderation factor $\gamma$ acts in the following way:

(1) When the sample is misclassified and $p_t$ is small (hard-to-classify sample), the adjustment parameter $(1-p_t)^\gamma$ is close to 1 and has almost no effect on the result of the loss function.

(2) When the sample $p_t$ is 1 (fully correctly classified sample), the adjustment parameter $(1-p_t)^\gamma$ is 0, and the loss of the fully correctly classified sample is 0, so that the model does not focus on such samples.

(3) When sample $p_t$ is 0.9 (easily classified correct samples) and $\gamma = 2$, the adjustment parameter $(1-p_t)^\gamma$ is 0.01 and the sample loss decreases to $\frac{1}{100}$ of the original, causing the model to focus less on such samples.

(4) When $\gamma = 0$, the Focal Loss Function becomes the Cross-Entropy Loss function.

In summary, the focal loss function is formulated as

$$FL(p_t) = -\alpha_t(1-p_t)^\gamma \log(p_t) \tag{9}$$

For the face recognition task, the large order of magnitude of the prior boxes causes the majority of samples to be background, which corresponds to the role of the $\alpha$ parameter in the Focal Loss function, and the introduction of $\gamma$ makes the model focus more on samples where no faces are recognized, which leads as a direct result to the background being recognized as faces over a large range with low confidence. Faces with low confidence can be excluded in the NMS stage. From the results, the introduction of the Focus Loss function does improve the model's detection of small faces, but the number of faces in the post-processing stage is large due to too many prior boxes and Oversensitive detection. Some "hard" images have thousands of faces (most of which can be eliminated), and this leads to NMS slowing down considerably.

Based on the above issues, this paper uses Cross-Entropy Loss for the first 245 training epochs, and Focal Loss Function for the last 5 training epochs so that the speed of NMS is almost unaffected by the improved accuracy of the model.

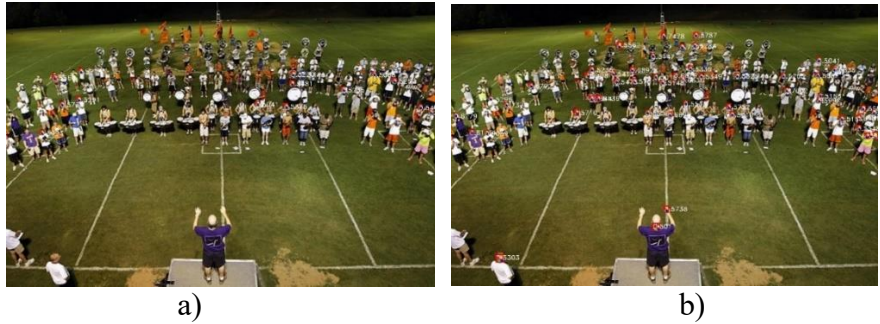

a)          b)

FIGURE 4. a) Results of the model without the improved method, b) Results of the model with the Focal Loss Function added.

As shown in Fig. 4, the recognition rate of small faces is much improved with the addition of the focal loss function. In Fig. a), only a small number of subtle faces are recognized in the normal model, but more small faces are recognized in Fig. b). At the same time, a false recognition phenomenon occurs, but in this case the recognition accuracy of Fig. b) is still higher than Fig. a). The combination of false recognition and the increased ability to recognize small faces validates the role of Focal Loss Function and illustrates that there are advantages and disadvantages to use this method, increasing recall and average accuracy while decreasing precision.

4. **Experiment and Results.** This section outlines the dataset and environment configurations used in the experiments, along with the accuracy comparisons and ablation experiments performed on the LAFD algorithm.

4.1. **Dataset.** We use the WIDERFACE dataset, which has a rich range of facial types that are difficult to recognize and has detailed information about each face, such as expression, omission, blur, etc.

4.2. **Environmental Configuration.** In this paper, the Pytorch framework was used for the training process.

The hardware and software environment for model training:

TABLE 2. Hardware and Software Configuration

| Operating System | Windows 10 |
|---|---|
| Python Version | 3.9 |
| Python Framework | Pytorch1.12 |
| CUDA | 11.3 |
| GPU | NVIDIA GTX 1660 Ti @6G |
| CPU | Intel Core i7-9750H @2.60GHz |
| RAM | 16 |
| ROM | 500 |

In this paper the LAFD model training model is set up with the following parameters:

TABLE 3. Training Parameter Configuration

| Batch Size | 5 |
|---|---|
| Image Size | 640*640 |
| Learning rate | 1e-3 |
| Training impulse | 0.9 |
| Epoch | 250 |
| optimization | SGD |
| Weight decay | 5e-4 |
| SGD gamma | 0.1 |

The parameters for the feature pyramid, prior box, and SSH settings in the model are

TABLE 4. Model Parameter Configuration

| Feature layer | output channel | feature map size | step | Prior Box size | Prior Box quantity | LeakyReLU |
|---|---|---|---|---|---|---|
| Stage1 | 40 | 80*80 | 8 | [16,32] | 80*80*2 | 0.1 |
| Stage2 | 112 | 40*40 | 16 | [64,128] | 40*40*2 | 0.1 |
| Stage3 | 160 | 20*20 | 32 | [256,512] | 20*20*2 | 0.1 |

Pre-processing of images before model detection and post-processing of model parameters:

TABLE 5. Pre-processing and post processing parameter configuration

| Image Length | Image Width | Detection Threshold | NMS Threshold |
|---|---|---|---|
| ≤ 1560 | ≤ 1200 | 0.5 | 0.4 |

Before testing the model, the images are proportionally expanded to a maximum length of 1560 or a maximum width of 1200. The model output was discarded if detection threshold was less than 0.5. If the IOU (Intersection over Union) value in the NMS is greater than 0.4, the face boxes with low detection accuracy were discarded.

4.3. **Accuracy Comparison.** Experiments comparing the LAFD algorithm with a number of widely used face detection algorithms:

TABLE 6. Comparison between LAFD and representative algorithms on WIDERFACE

| Model | Easy (AP) | Medium (AP) | Hard (AP) | Model Size (MB) |
|---|---|---|---|---|
| V-J | 0.412 | 0.333 | 0.137 | — |
| DPM [16] | 0.690 | 0.448 | 0.201 | — |
| Faceness-Net [7] | 0.704 | 0.573 | 0.273 | — |
| ScaleFace [17] | 0.868 | 0.867 | 0.772 | — |
| CMS-RCNN [18] | 0.899 | 0.874 | 0.624 | — |
| Retinaface | 0.907 | 0.882 | 0.738 | 1.7 |
| LFFD | 0.910 | 0.881 | 0.780 | 9 |
| SSD | 0.928 | 0.916 | 0.832 | 95 |
| SSH | 0.931 | 0.921 | 0.845 | — |
| Face-R-CNN [19] | 0.937 | 0.921 | 0.831 | — |
| FANet [20] | 0.945 | 0.947 | 0.895 | 83 |
| yolov7-tiny [8] | 0.947 | 0.926 | 0.821 | 11.8 |
| TinaFace [21] | 0.963 | 0.957 | 0.930 | 144 |
| LAFD (ours) | 0.943 | 0.926 | 0.862 | 10.2 |

As shown in Table 6, the LAFD model achieves the average accuracy of 94.3%, 92.6% and 86.2% on the three validation subsets of WIDERFACE, which is 3.6%, 4.4% and 12.4% higher than Retinaface, respectively. For V-J and DPM, traditional detection methods that do not use convolutional neural networks, the accuracy is low in the three subsets due to the lack of methods for effective feature extraction; for Faceness-Net, the lack of methods for multi-size feature extraction of faces, the average accuracy of its "hard" subset is low; for ScaleFace, no attention mechanism is introduced in the deep network, resulting in the model accuracy is low compared to other models; for SSH and SSD single-stage face detectors, both use feature pyramids for multi-scale face feature extraction with better results; for FANet and TinaFace, their models are too large in size and less likely to be used in embedded situations which requires a lightweight model. Yolov7-tiny, a lightweight model introduced in July 2022, performs similarly to LAFD on the WIDERFACE "easy" and "medium" validation sets, but the average accuracy of LAFD on "hard" validation set is 4.1% higher than yolov7-tiny, and the model size of LAFD is 1.6MB smaller.

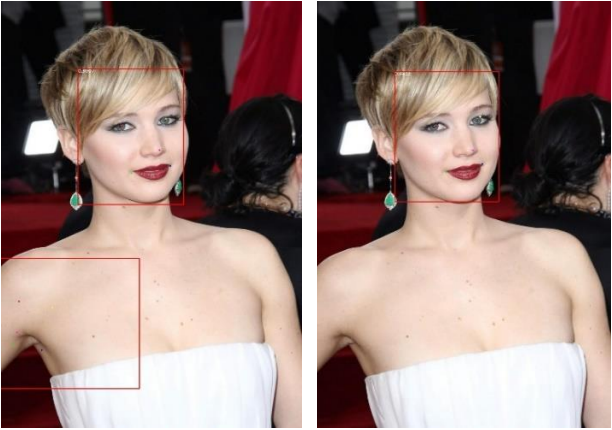

a) Retinaface results for female portraits   b) LAFD results for female portraits

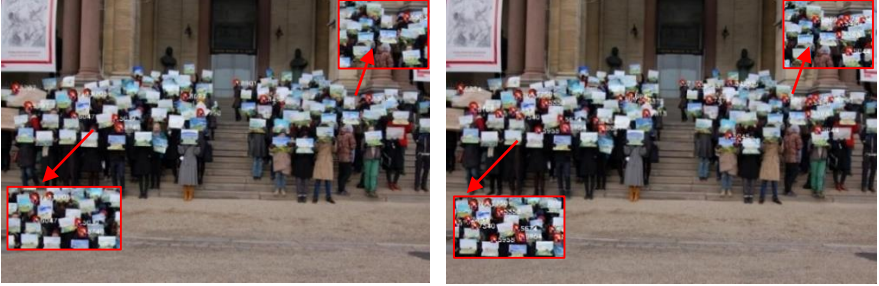

c) Retinaface results for group shots   d) LAFD Results for group shots

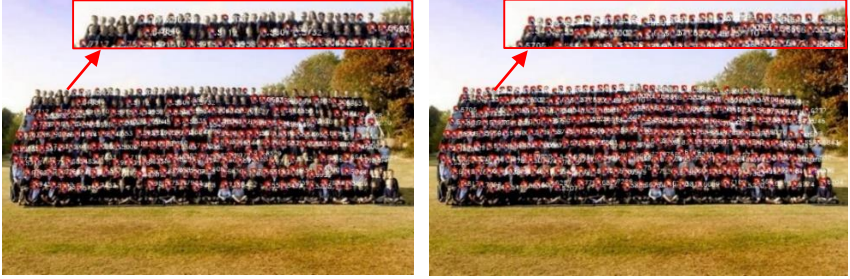

e) Retinaface results for the graduation photo   f) LAFD results for the graduation photo

FIGURE 5. Experimental results of Retinaface (left panel) versus LAFD (right panel).

Fig. 5 shows the comparison between the experimental results of Retinaface and LAFD. In Fig. a) and b), Retinaface misidentifies the face in the lower left corner of the original image, while the LAFD model detects the face normally. In Fig. c) and d), Retinaface is unable to fully recognize occluded faces while LAFD performs better on such images. In Fig. e) and f), Retinaface is unable to accurately identify small faces in the distance while LAFD recognizes almost all faces. The above comparison of the two models in the WIDERFACE validation set provides ample evidence that LAFD is not only numerically but also practically more accurate than Retinaface.

4.4. **Ablation Experiment.** To explain the improvement of each added module in the deep convolutional network, this paper conducts ablation experiments. The models are trained and tested on the WIDERFACE dataset in four modules: new backbone network, Focal Loss Function, deformable convolution network, and resizing of the test images. The results of the ablation experiments are presented below.

TABLE 7. Ablation Experiment

| Mobile NetV1 | Mobile NetV3 | MobileNet -Improved | Focal Loss | DCN | Resize | Easy (AP) | Medium (AP) | Hard (AP) |
|---|---|---|---|---|---|---|---|---|
| 1 | 0 | 0 | 0 | 0 | 0 | 0.907 | 0.882 | 0.739 |
| 0 | 1 | 0 | 0 | 0 | 0 | 0.929 | 0.907 | 0.784 |
| 0 | 0 | 1 | 0 | 0 | 0 | 0.927 | 0.905 | 0.797 |
| 0 | 0 | 1 | 1 | 0 | 0 | 0.927 | 0.906 | 0.805 |
| 0 | 0 | 1 | 1 | 1 | 0 | 0.927 | 0.904 | 0.800 |
| 0 | 0 | 1 | 0 | 1 | 0 | 0.941 | 0.922 | 0.821 |
| 0 | 0 | 1 | 0 | 1 | 1 | 0.943 | 0.926 | 0.862 |

Note: Resize means that the input image is pre-processed and scaled to 1560px in length or 1200px in width. Number 0 in the cell indicates that this method was not used and number 1 indicates that this method was used.

From Table 7, it is concluded that the new backbone network, the replacement of loss function, the deformable convolution and resizing input image are all effective for Retinaface. The introduction of DCN and Focal Loss is contradictory and using the Focal Loss function with deformable convolution leads to a decrease in accuracy. This paper therefore selects the DCN that provides the higher accuracy improvement of the two. The model uses a modified MobilenetV3 backbone network with DCN replacing the 3x3 convolution in the SSH layer, and the results are improved by 3.3%, 3.9%, and 8.5% in three subsets compared to Retinaface. Scaling the input images to 1560px in length or 1200px in width improved the accuracy of the model by 3.3%, 4.1%, and 12.4% in three subsets respectively compared to Retinaface.

5. **Conclusions.** RetinaFace is a single-stage lightweight face detection network. This paper improves its backbone network MobileNet-V3 by fine-tuning the use of the SE attention mechanism, the channel expansion multiplier of the Inverted Residuals Block and the size of the convolution kernel to make it more suitable for face detection. The Deformable

Convolution Network is used to replace the original convolution in the SSH layer; the original Cross-Entropy loss is replaced by the Focus Loss function; the image is pre-processed by expanding the input image to 1560px in length and width or 1200px in width in equal proportion.

In future improvements, the author will try to use GIOU, DIOU and other 2D loss functions instead of the loss function Smooth-L1, explore the composite relationship between the Focal Loss function and DCN and continue to adjust the parameters of the MobileNetV3 backbone network.

**Acknowledgment.** This work is supported by Natural Science Foundation of Guangdong Province (2023A1515012894), Key R&D Project of Guangzhou Science and Technology Plan(2023B01J0002). The authors also gratefully acknowledge the helpful comments and suggestions of the reviewers, which have improved the presentation.